\documentclass[runningheads]{llncs}

\usepackage{eccv}

\usepackage{eccvabbrv}

\usepackage{graphicx}
\usepackage{booktabs}
\usepackage{amsmath, cite, amssymb,amsfonts, algorithmic, textcomp, multirow, subcaption, comment, balance}

\usepackage[accsupp]{axessibility}  % Improves PDF readability for those with disabilities.

\usepackage[pagebackref,breaklinks,colorlinks,citecolor=eccvblue]{hyperref}
\usepackage{hyperref}

\usepackage{orcidlink}

\begin{document}

\title{Braided Vision Transformer for Stroke Detection in Multi-view Retinal Fundus Imaging} 

\titlerunning{BViT for Stroke Detection}

\author{Aysen Degerli\orcidlink{0000-0002-9478-033X} \and
Mika Hilvo\orcidlink{0000-0003-0663-5843}}

\authorrunning{A. Degerli and M. Hilvo}

\institute{VTT Technical Research Centre of Finland, Tampere and Espoo, Finland \\
\email{name.surname@vtt.fi}}

\maketitle

\begin{abstract}
  Stroke remains a leading cause of mortality and morbidity worldwide, emphasizing the importance of its accurate and immediate assessment. Retinal fundus imaging has emerged as a promising modality for stroke assessment, as the retina reflects cerebrovascular and neurological risk factors. Contrary to conventional neuroimaging techniques, retinal fundus imaging offers a non-invasive, cost-effective, and portable alternative for rapid screening. This paper explores the feasibility of retinal fundus imaging for stroke and transient ischemic attack (TIA) detection using macula-centric and optic nerve head-centric views captured from both eyes. Our study introduces, to the best of our knowledge, the first vision transformer model for retinal fundus imaging in stroke assessment, offering a novel approach for capturing retinal patterns. Thereby, we propose the Braided Vision Transformer (BViT) model, which extracts representative features from the given multi-view images while simultaneously capturing inter‑view relationships across both eyes, enabling a more informative understanding of retinal biomarkers associated with cerebrovascular events. Experiments conducted on our collected Stroke‑Data dataset demonstrate that BViT achieves an AUC score of 0.75 for stroke detection, outperforming regular vision transformers.
  \keywords{Vision Transformers \and Multi-view Learning \and Retinal Fundus Imaging}
\end{abstract}

\section{Introduction}
World Health Organization (WHO) recognizes stroke globally as the second leading cause of mortality, with approximately $6.5$ million deaths reported each year by the World Stroke Organization (WSO) \cite{campbell2019ischaemic, feigin2022world}. Cerebrovascular diseases affect the blood vessels that supply blood to the brain by an occluded or ruptured artery, which results in stroke or transient ischemic attack (TIA) \cite{ABSHER2002733}. According to WHO, a stroke is defined by symptoms lasting $24$ hours or more, potentially resulting in death \cite{hatano1976experience}, while a TIA, often referred to as a mini-stroke, causes similar symptoms that resolve within a day without lasting damage \cite{amarenco2020transient}. Importantly, TIA provides a critical early warning sign, indicating a $20-25\%$ risk of a subsequent stroke \cite{amarenco2020transient}. Hence, the development of stroke risk assessment strategies is essential to mitigate the substantial global burden of cerebrovascular diseases.  

Neuroimaging techniques are essential for initial stroke identification, with Computed Tomography (CT) and Magnetic Resonance Imaging (MRI) commonly employed to monitor cerebral infarction \cite{birenbaum2011imaging}. Despite these imaging techniques being the gold standard for stroke evaluation, their clinical usage is constrained by several factors, including long acquisition time, high operational costs, and, in the case of CT imaging, exposure to ionizing radiation. Hence, the aforementioned limitations motivate a growing interest in alternative imaging techniques in the biomedical field that may provide a rapid, robust, and safer assessment of cerebrovascular events \cite{10.1159/000362719, muir2006imaging}. Retinal fundus imaging, acquired by a fundus camera, is considered a potential imaging technique for assessing stroke risk, as retinal microvasculature alterations expose cerebrovascular health \cite{CHEUNG201789, rim2020retinal}. Accordingly, retinal arterial abnormalities associated with stroke risk include retinal venular widening, cotton wool spots, increased tortuosity, and arterial narrowing \cite{CHEUNG201789, henderson2011hypertension, rim2020retinal}. Consequently, retinal fundus imaging offers a non-invasive, portable, and cheaper option for stroke risk assessment compared to conventional CT and MRI.

\begin{figure*}[t!]
    \centering
    \includegraphics[width=.95\linewidth]{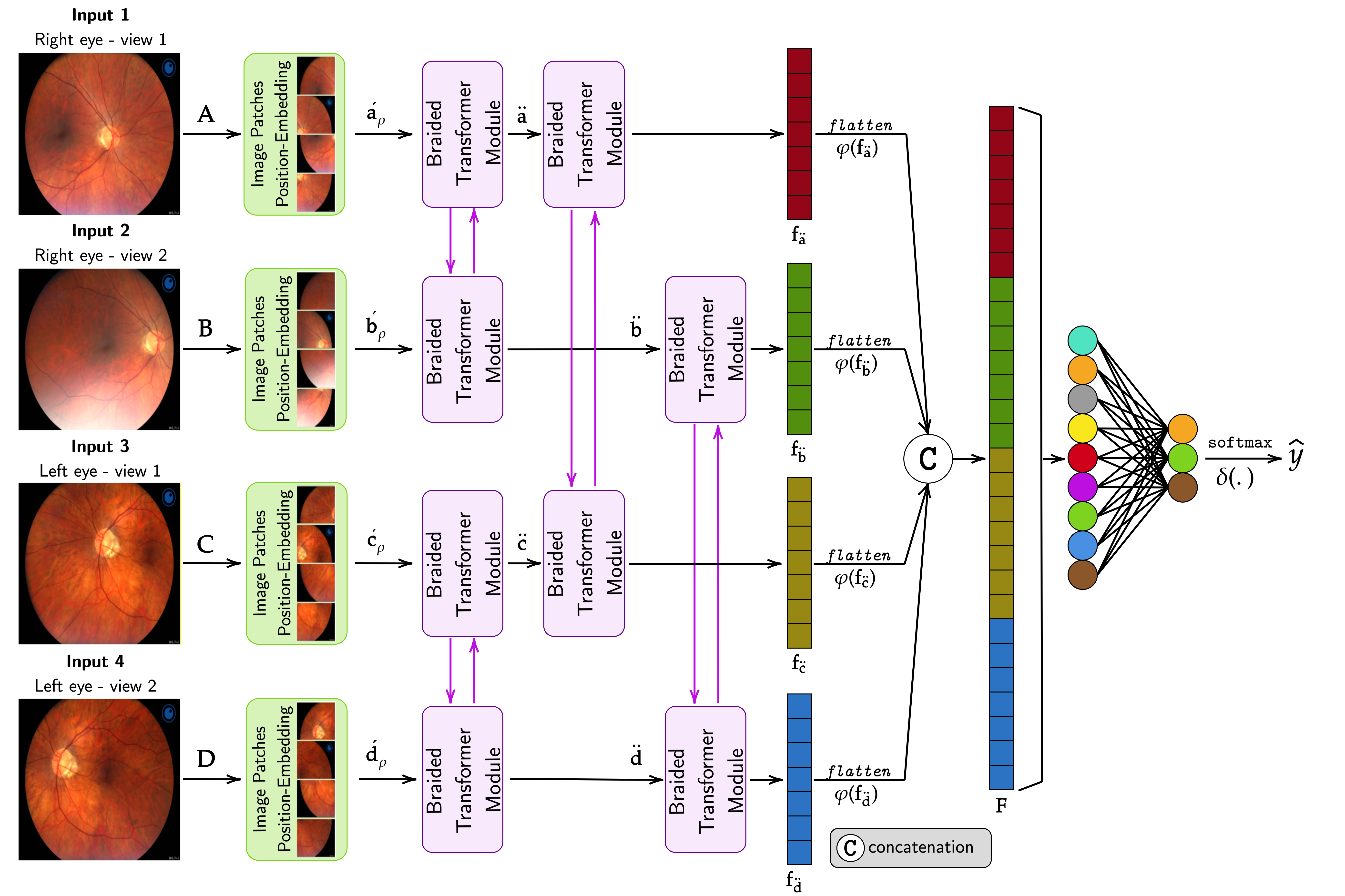}
    \caption{The illustration of the proposed Braided Vision Transformer model structure that is designed to detect stroke from macula-centric and optic nerve head-centric retinal fundus images collected from both the left and right eye.
    \label{fig:framework}}
\end{figure*}

In this study, we propose a novel transformer architecture, Braided Vision Transformer (BViT) designed to analyze macula-centric and optic nerve head-centric view retinal fundus images collected from both eyes, as depicted in Fig. \ref{fig:framework}. To the best of our knowledge, Vision Transformers (ViTs) have not yet been explored for retinal fundus images in the context of stroke assessment, despite their state-of-the-art performance in other recent computer vision tasks. This work, therefore, introduces the first transformer-based architecture for retinal fundus imaging for stroke detection. BViT captures the inter-view relationship across the retinal field by modeling the input channels in a braided structural form that allows the model to leverage multi-view information from each eye and thus extract a more informative understanding of ocular features. Hence, the proposed braided structure effectively mimics the clinical evaluation practices for stroke assessment.

\section{Related Work}
Deep learning is widely employed in the interpretation of medical imaging data, and thus many studies \cite{pachade2022detection, Lim_Lim_Xu_Ting_Wong_Lee_Hsu_2019, coronado2021towards, diagnostics12071714} have utilized it also for stroke assessment with retinal fundus images. In this context, deep learning has been used in feature extraction \cite{pachade2022detection}, to develop state-of-the-art Convolutional Neural Networks (CNNs) \cite{Lim_Lim_Xu_Ting_Wong_Lee_Hsu_2019}, and to perform transfer learning with pre-trained deep networks \cite{coronado2021towards, diagnostics12071714}. Although several promising solutions have been presented for stroke assessment, studies have utilized only single-view retinal fundus images acquired from one eye, which may not capture the entire inter-eye vascular variation observed in routine clinical evaluations. Moreover, studies have focused only on ischemic strokes, while TIAs have not been investigated. 

Multi-view learning (MVL) has growing interest across many fields, including biomedical research, due to the increase in diverse data types \cite{YAN2021106}. Conventionally, deep MVL is employed by CNNs that map data from different views into a common shared space representing high-level features for further analysis. In the literature, only one study \cite{degerli2025advancedassessmentstrokeretinal} has proposed multi-view retinal fundus imaging assessment for stroke detection using CNNs. On the other hand, transformer models have progressively replaced CNNs due to their outstanding performance. Hence, there is an urge to employ multi-view learning paradigms in the transformer-based architectures. Following the introduction of the transformer model by Vaswani et al. \cite{NIPS2017_3f5ee243} that was initially proposed for machine translation, subsequent development of the vision transformer (ViT) \cite{dosovitskiy2021an} marked a significant shift in visual recognition tasks. Further developments in several studies \cite{wu2022multimodal, zhou2021deep, lei2022multi} have deployed MVL in ViT models by proposing a mutual attention transformer module to leverage information from different modalities.

\section{Methodology}
In this section, we define the problem and present the network architecture and training of the proposed BViT model.
 
\subsection{Problem Formulation}
In this study, we use dual-view retinal fundus images, referring to macula-centric and optic nerve head-centric views, acquired from both eyes, to detect the presence of stroke or TIA. The proposed BViT model is designed to distinguish stroke together with TIA from healthy controls, considered as a binary classification framework. In addition, we formulate a multi-class task, where BViT is trained to discriminate stroke, TIA, and healthy controls. Our primary emphasis relies on the binary classification task, as stroke and TIA often resemble each other with similar early-stage symptoms, making timely and accurate detection critically important.

A set of multi-view retinal fundus images are denoted as $\mathbf{M} =\{\mathbf{I}_{r,{v_1}}, \mathbf{I}_{r,{v_2}},\allowbreak \mathbf{I}_{l,{v_1}}, \mathbf{I}_{l,{v_2}}\}$, where $\mathbf{I} \in \mathbb{R}^{h \times w \times c}$  denotes an RGB color image with height of $h$, width of $w$, and $c=3$ number of channels, optic nerve head-centric, $v_1$ and macula-centric, $v_2$ views of left, $l$ and right, $r$ eyes. The proposed BViT, referred as $\Theta$, maps the given multi-view image set $\mathbf{M}$ and ground truth labels $y \in \{0,1,2\}$ to their predictions $\hat{y}$ as $\hat{y} \leftarrow \Theta(\mathbf{M}, y)$ and trained over Stroke-Data dataset, $\mathbf{D}=\{\mathbf{M}_s, \mathbf{y}_s\}_{s=1}^S$ with $S$ number of samples.

\subsection{Model Structure}
The proposed BViT model consists of four input channels for the given left and right eye images from dual-view retinal fundus images, each attached to a position embedding block. Let us denote a single image $\mathbf{A}$ and its following patches $\mathbf{a}_\rho$. The block constructs a sequence of flattened patches by reshaping $\mathbf{A}$ into $\mathbf{a}_\rho \in {\mathbb{R}^{t \times (p^2 c)}}$, where the patch size of the reshaped image is $(p \times p)$, and $t=hw/p^2$ is the number of patches resulted from $\mathbf{A}$. The linear projection of the patches is obtained by flattening each image patch defined as $\acute{\mathbf{a}}_\rho=\varphi(\mathbf{a}_\rho)$. Then, positional embedding is applied with a feed-forward layer, which is added to the flattened patch, $\acute{\mathbf{a}}_\rho$ via summation. 

\begin{figure}[t!]
    \centering
    \includegraphics[width=\linewidth]{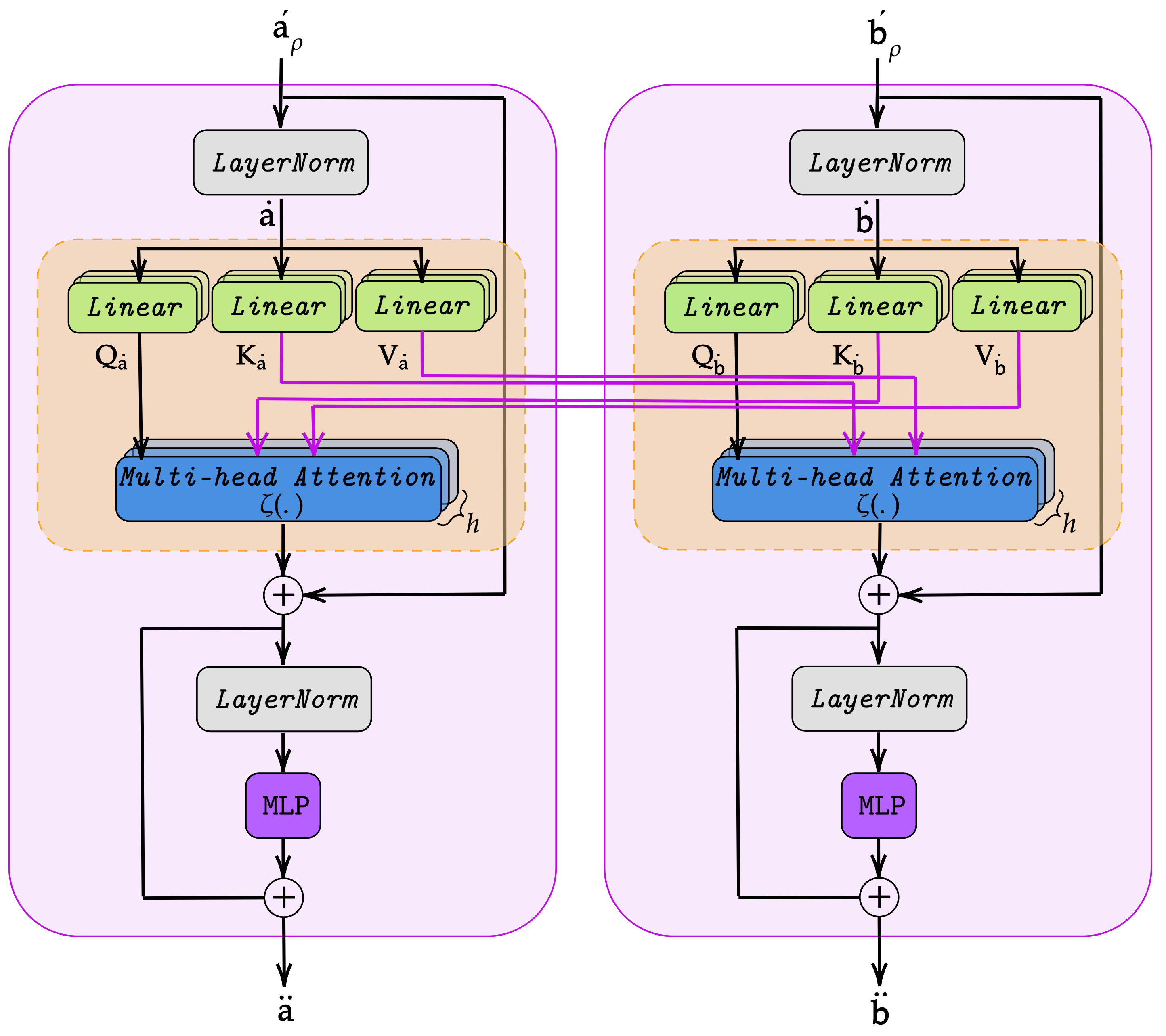}
    \caption{The structure of the proposed braided transformer module.}
    \label{fig:braided_module}
\end{figure}

The position-embedded image patches are the inputs of the braided transformer modules as shown in Fig. \ref{fig:braided_module}. The patches are further processed with layer normalization (\textit{LayerNorm}) layers, $\dot{\mathbf{a}} \in \mathbb{R}^{N \times d}$. An attention function of the given query $\mathbf{Q}$, key $\mathbf{K}$, and value $\mathbf{V}$ is defined as
\begin{equation}
\zeta(\mathbf{Q,K,V})=\delta\left(\frac{\mathbf{QK}^T}{\sqrt{d_k}}\right)\mathbf{V}
\end{equation}
\noindent where and $\delta(.)$ is the \textit{softmax} function and $d_k$ is the scaling factor, and $\zeta(.)$ directs the attention of the model towards the most relevant parts of the given input by weighting the sum of values. In the vanilla vision transformer block proposed by Dosovitskiy et al. \cite{dosovitskiy2021an}, the multi-head attention module performs in parallel multiple attention functions as
\begin{equation}
    \begin{aligned}
        \mu(\mathbf{Q,K,V})&=\Omega(h_1, h_2, ...,h_m)\mathbf{W}^O, \quad \text{where} \\
        h_i=\zeta(\mathbf{Q}, \mathbf{K}, \mathbf{V})&=\zeta(\mathbf{QW}_i^\mathbf{Q}, \mathbf{KW}_i^\mathbf{K}, \mathbf{VW}_i^\mathbf{V}), \\
    \end{aligned}
\end{equation}
\noindent $m$ is the number of heads, $\Omega$ is the concatenation operation, and $\mathbf{W}_i$ are the projection matrices. In mutual transformer module, Wu et al. \cite{wu2022multimodal} proposed to connect two given embeddings of input modalities $\mathbf{A, B} \in \mathbb{R}^{h \times w \times c}$, where $\mathbf{\dot{a},\dot{b}} \in \mathbb{R}^{N \times d}$ are the embeddings after layer normalization with their corresponding queries, keys, and values $\mathbf{Q}_\mathbf{\dot{a}},\mathbf{K}_\mathbf{\dot{a}},\mathbf{V}_\mathbf{\dot{a}} \in \mathbb{R}^{N \times d}$ and $\mathbf{Q}_\mathbf{\dot{b}},\mathbf{K}_\mathbf{\dot{b}},\mathbf{V}_\mathbf{\dot{b}} \in \mathbb{R}^{N \times d}$. The attention for the mutual transformer module is described below: 
\begin{equation}
    \begin{aligned}
        \mu(\mathbf{Q}_{\mathbf{\dot{a}}},\mathbf{K}_{\mathbf{\dot{b}}},\mathbf{V}_{\mathbf{\dot{b}}})&=\Omega(h_{\mathbf{\dot{a}},1}, h_{\mathbf{\dot{a}},2}, ...,h_{\mathbf{\dot{a}},m})\mathbf{W}^O, \\
        \text{where } \quad
        h_{\mathbf{\dot{a}},j} &= \delta\left(\frac{\mathbf{Q}_{\mathbf{\dot{a}},j}\mathbf{K}_{\mathbf{\dot{b}},j}^T}{\sqrt{d_k}}\right)\mathbf{V}_{\mathbf{\dot{b}},j}, \\
        \mu(\mathbf{Q}_\mathbf{\dot{b}},\mathbf{K}_\mathbf{\dot{a}},\mathbf{V}_\mathbf{\dot{a}})&=\Omega(h_{\mathbf{\dot{b}},1}, h_{\mathbf{\dot{b}},2}, ...,h_{\mathbf{\dot{b}},m})\mathbf{W}^O, \\
        \text{where } \quad
        h_{\mathbf{\dot{b}},j} &= \delta\left(\frac{\mathbf{Q}_{\mathbf{\dot{b}},j}\mathbf{K}_{\mathbf{\dot{a}},j}^T}{\sqrt{d_k}}\right)\mathbf{V}_{\mathbf{\dot{a}},j},
    \end{aligned}
    \label{eq:mutual_MHA}
\end{equation}
\noindent with $j$th head for $\mathbf{\dot{a}}$ and $\mathbf{\dot{b}}$. In the mutual transformer module in \cite{wu2022multimodal}, there are two multi-head attention blocks, each taking the key and value from the other modality. In addition, the outputs of the multi-head attention blocks are added together for each modality.

In BViT, the braided transformer module connects the four embeddings of the input images $\mathbf{A, B, C, D} \in \mathbb{R}^{h \times w \times c}$ corresponding to $\mathbf{I}_{r, v_1}, \mathbf{I}_{r, v_2}, \mathbf{I}_{l, v_1}, \mathbf{I}_{l, v_2}$ with a braided structure as depicted in Fig. \ref{fig:framework}. Accordingly, BViT has four braided transformer modules attached to the image embeddings, $\mathbf{\dot{a}, \dot{b}, \dot{c}, \dot{d}} \in \mathbb{R}^{N \times d}$ with their corresponding queries, keys, and values $\{\mathbf{Q}_{\mathbf{\dot{a}}}, \mathbf{K}_{\mathbf{\dot{a}}}, \mathbf{V}_{\mathbf{\dot{a}}}\}$, \allowbreak $\{\mathbf{Q}_{\mathbf{\dot{b}}}, \mathbf{K}_{\mathbf{\dot{b}}}, \allowbreak \mathbf{V}_{\mathbf{\dot{b}}}\}$, $\{\mathbf{Q}_{\mathbf{\dot{c}}}, \mathbf{K}_{\mathbf{\dot{c}}}, \mathbf{V}_{\mathbf{\dot{c}}}\}$ and $\{\mathbf{Q}_{\mathbf{\dot{d}}}, \mathbf{K}_{\mathbf{\dot{d}}}, \mathbf{V}_{\mathbf{\dot{d}}}\} \in \mathbb{R}^{N \times d}$. In addition to the multi-head attention of the image embeddings $\mathbf{\dot{a}}$ and $\mathbf{\dot{b}}$ described in Eq. (\ref{eq:mutual_MHA}), the braided transformer module processes the embeddings $\mathbf{\dot{c}}$ and $\mathbf{\dot{d}}$  as follows:
\begin{equation}
    \begin{aligned}
        \mu(\mathbf{Q}_\mathbf{\dot{c}},\mathbf{K}_\mathbf{\dot{d}},\mathbf{V}_\mathbf{\dot{d}})&=\Omega(h_{\mathbf{\dot{c}},1}, h_{\mathbf{\dot{c}},2}, ...,h_{\mathbf{\dot{c}},m})\mathbf{W}^O, \\
        \text{where } \quad
        h_{\mathbf{\dot{c}},j} &= \delta\left(\frac{\mathbf{Q}_{\mathbf{\dot{c}},j}\mathbf{K}_{\mathbf{\dot{d}},j}^T}{\sqrt{d_k}}\right)\mathbf{V}_{\mathbf{\dot{d}},j}, \\
        \mu(\mathbf{Q}_\mathbf{\dot{d}},\mathbf{K}_\mathbf{\dot{c}},\mathbf{V}_\mathbf{\dot{c}})&=\Omega(h_{\mathbf{\dot{d}},1}, h_{\mathbf{\dot{d}},2}, ...,h_{\mathbf{\dot{d}},m})\mathbf{W}^O, \\
        \text{where } \quad
        h_{\mathbf{\dot{d}},j} &= \delta\left(\frac{\mathbf{Q}_{\mathbf{\dot{d}},j}\mathbf{K}_{\mathbf{\dot{c}},j}^T}{\sqrt{d_k}}\right)\mathbf{V}_{\mathbf{\dot{c}},j}.
    \end{aligned}
\end{equation}
\noindent Following the multi-head attention in the braided transformer modules, its output is added to $\mathbf{\acute{a}}_{\rho}$. Next, \textit{LayerNorm} and multi-layer perceptron (MLP) layers are attached before the final addition layer, as depicted in Fig. \ref{fig:braided_module}. 

To complete the braided structure, an additional block of braided transformer modules is attached to the outputs of the first modules, $\mathbf{\ddot{a}, \ddot{b}, \ddot{c}, \ddot{d}}$ below:
\begin{equation}
    \begin{aligned}
        \mu(\mathbf{Q}_\mathbf{\ddot{a}},\mathbf{K}_\mathbf{\ddot{c}},\mathbf{V}_\mathbf{\ddot{c}})&=\Omega(h_{\mathbf{\ddot{a}},1}, h_{\mathbf{\ddot{a}},2}, ...,h_{\mathbf{\ddot{a}},m})\mathbf{W}^O, \\
        \mu(\mathbf{Q}_\mathbf{\ddot{b}},\mathbf{K}_\mathbf{\ddot{d}},\mathbf{V}_\mathbf{\ddot{d}})&=\Omega(h_{\mathbf{\ddot{b}},1}, h_{\mathbf{\ddot{b}},2}, ...,h_{\mathbf{\ddot{b}},m})\mathbf{W}^O, \\
        \mu(\mathbf{Q}_\mathbf{\ddot{c}},\mathbf{K}_\mathbf{\ddot{a}},\mathbf{V}_\mathbf{\ddot{a}})&=\Omega(h_{\mathbf{\ddot{c}},1}, h_{\mathbf{\ddot{c}},2}, ...,h_{\mathbf{\ddot{c}},m})\mathbf{W}^O, \\
        \mu(\mathbf{Q}_\mathbf{\ddot{d}},\mathbf{K}_\mathbf{\ddot{b}},\mathbf{V}_\mathbf{\ddot{b}})&=\Omega(h_{\mathbf{\ddot{d}},1}, h_{\mathbf{\ddot{d}},2}, ...,h_{\mathbf{\ddot{d}},m})\mathbf{W}^O.
    \end{aligned}
\end{equation}
\noindent The output of each braided transformer module is then flattened and concatenated into a feature vector defined as $\mathbf{F}=[\mathbf{f}_{\ddot{a}}, \mathbf{f}_{\ddot{b}}, \mathbf{f}_{\ddot{c}}, \mathbf{f}_{\ddot{d}}]$,  $\mathbf{f} \in \mathbb{R}^D$, where $D$ is the feature dimension after flattening. Then, a fully connected layer with rectified linear unit (\textit{ReLU}) activation function and $D/6$ number of neurons is attached to $\mathbf{F}$. For the output layer, a fully connected layer with \textit{softmax} activation function with a neuron size dependent on the number of classes, where in binary classification its size is set to $2-$neurons, whereas for multi-class classification $3-$neurons are used.

\subsection{Model Training}
\textbf{Data preparation.} In this study, we used the Stroke-Data dataset, which is from study \cite{degerli2025advancedassessmentstrokeretinal}, collected at the Oulu Kuopio University Hospitals in Finland between $2021$ and $2022$, consisting of $220$ participants, including $26$ TIA and $73$ stroke patients, and $121$ healthy controls. In total, $802$ retinal fundus images were acquired with a set of four images collected per participant. The inequality in the number of images per eye view in the cohort arises from missing data during acquisition, mainly caused by low‑quality images or incomplete data collection for some patients.  When a specific view is missing, we replace it with horizontally flipping the corresponding view from the contralateral eye, if available. Retinal fundus images were resized to $224 \times 224$ pixels, and patient‑wise separated into training and test sets with a $4/1$ ratio. During model training, we adapted a stratified $5$-fold cross-validation scheme.

\textbf{Data augmentation.} Deep learning models trained on small datasets are at risk of overfitting, as they may memorize individual training samples rather than learn clinically meaningful pathological characteristics. Data augmentation addresses this limitation by artificially increasing data diversity, thereby improving model generalization and robustness. Through the introduction of controlled spatial transformations, the model learns to focus on disease-related vascular features while becoming less sensitive to irrelevant variations in image appearance. To reflect realistic acquisition conditions without compromising biological plausibility, only minor rotations and translations were applied. These augmentations help the model develop invariance to common imaging variations and artifacts arising from fundus camera positioning, patient head orientation, misalignment during acquisition, changes in scale, and illumination differences. To strengthen the model's robustness, data augmentation was applied extensively to the training set by augmenting per-class dual‑view left and right eye images to $1000$ samples, resulting in a total of $12$K images. Data augmentation included horizontal and vertical shifts, flipping, random shear transformation, random rotations of images in $0-180$ degrees, and zooming within a range of $10\%$.

\textbf{Model parameters. } The patch size of $112$ was used for the image patching before the transformer modules. In the BViT model, we used $2$ blocks of braided transformer modules sequentially for each input channel. In these modules, the number of heads for the multi-head attention was set to $3$, and the neuron size of $32$ was used for the MLP layer. The model was trained with the categorical cross-entropy loss function and the Adam optimizer. The parameter values of $10^{-5}$, $300$, and $32$ were used for the learning rate, epochs, and batch size, respectively. During training, the categorical accuracy of the model was monitored, the weights of the model with the highest score were saved, and early stopping with $5$ epochs of patience was applied. Lastly, the model was trained on an NVidia® Amper A$100$ GPU card with $40$ GB of memory. 

\textbf{Performance metrics.} The metrics that determine the performance are calculated from the accumulated confusion matrix obtained from folds, which is formed by true positive (TP), true negative (TN), false positive (FP), and false negative (FN) elements. For the model predictions, we apply a cut-off value of $0.5$ to calculate sensitivity ($\frac{\text{TP}}{\text{TP}+\text{FN}}$), specificity ($\frac{\text{TN}}{\text{TN}+\text{FP}}$), precision ($\frac{\text{TP}}{\text{TP}+\text{FP}}$), F1-Score by the harmonic mean of sensitivity and precision, and accuracy ($\frac{\text{TP}+\text{TN}}{\text{TP}+\text{TN}+\text{FP}+\text{FN}}$). In addition, the area under the curve (AUC) is measured from the Receiver Operating Characteristic (ROC) plot, which shows the sensitivity and $1-$specificity values by changing the cut-off value for the predictions.

\begin{table*}[t!]
\centering
\caption{Average stroke and TIA detection performance results of vision transformers with different transformer modules computed from 5-folds, where the highest scores are highlighted in \textbf{bold}.}
\resizebox{.95\linewidth}{!}{
\begin{tabular}{lccccccc}
\toprule
\multicolumn{1}{l}{Model} & \multicolumn{1}{c}{\begin{tabular}[c]{@{}c@{}}Transformer\\ Module\end{tabular}} & \multicolumn{1}{c}{Sensitivity} & \multicolumn{1}{c}{Specificity} & \multicolumn{1}{c}{Precision} & \multicolumn{1}{c}{F1-Score} & \multicolumn{1}{c}{AUC} & \multicolumn{1}{c}{Accuracy} \\ 
\midrule
 B-ViT & Braided (ours) & $0.647$ & $0.727$& $\textbf{0.660}$ & $\textbf{0.653}$ & $\textbf{0.752}$ & $\textbf{0.691}$ \\

 ViT & Vanilla \cite{dosovitskiy2021an} & $0.576$ & $\textbf{0.744}$ & $0.648$ & $0.610$ & $0.729$ & $0.668$ \\
 
 B-ViT & Mutual \cite{wu2022multimodal} & $0.576$ & $0.702$ & $0.613$ & $0.594$ & $0.717$ & $0.645$ \\

 ViT & Mutual \cite{wu2022multimodal} & $0.636$ & $0.727$ & $0.656$ & $0.646$ & $0.729$ & $0.686$ \\
 
 MVS-Net-v1 \cite{degerli2025advancedassessmentstrokeretinal} & - & $\textbf{0.667}$ & $0.661$ & $0.617$ & $0.641$ & $0.707$ & $0.664$  \\

 MVS-Net-v2 \cite{degerli2025advancedassessmentstrokeretinal} & - & $0.606$ & $0.678$ & $0.606$ & $0.606$ & $0.710$ & $0.646$  \\

\bottomrule 
\end{tabular}}
\label{tab:experiments_1}
\end{table*}

\section{Results}
In this section, we report the experimental results and compare the performance of BViT against the baselines ViT vanilla \cite{dosovitskiy2021an}, ViT mutual \cite{wu2022multimodal}, and BViT mutual \cite{wu2022multimodal}. In order to have a fair comparison, all models were configured using the same model parameters as in BViT, including the patch size, number of transformer blocks, MLP hidden dimensions, and number of heads in the multi-head attention. In ViT vanilla \cite{dosovitskiy2021an}, there is no connection between the input channels, whereas the mutual transformer module \cite{wu2022multimodal} introduces partial cross-channel interactions by connecting only the top two and bottom two input channels separately. In contrast, the proposed BViT model with the braided transformer module establishes an intersectional connectivity among all input channels, enhancing cross-channel integration.

\begin{figure}[t!]
    \centering
    \includegraphics[width=0.8\linewidth]{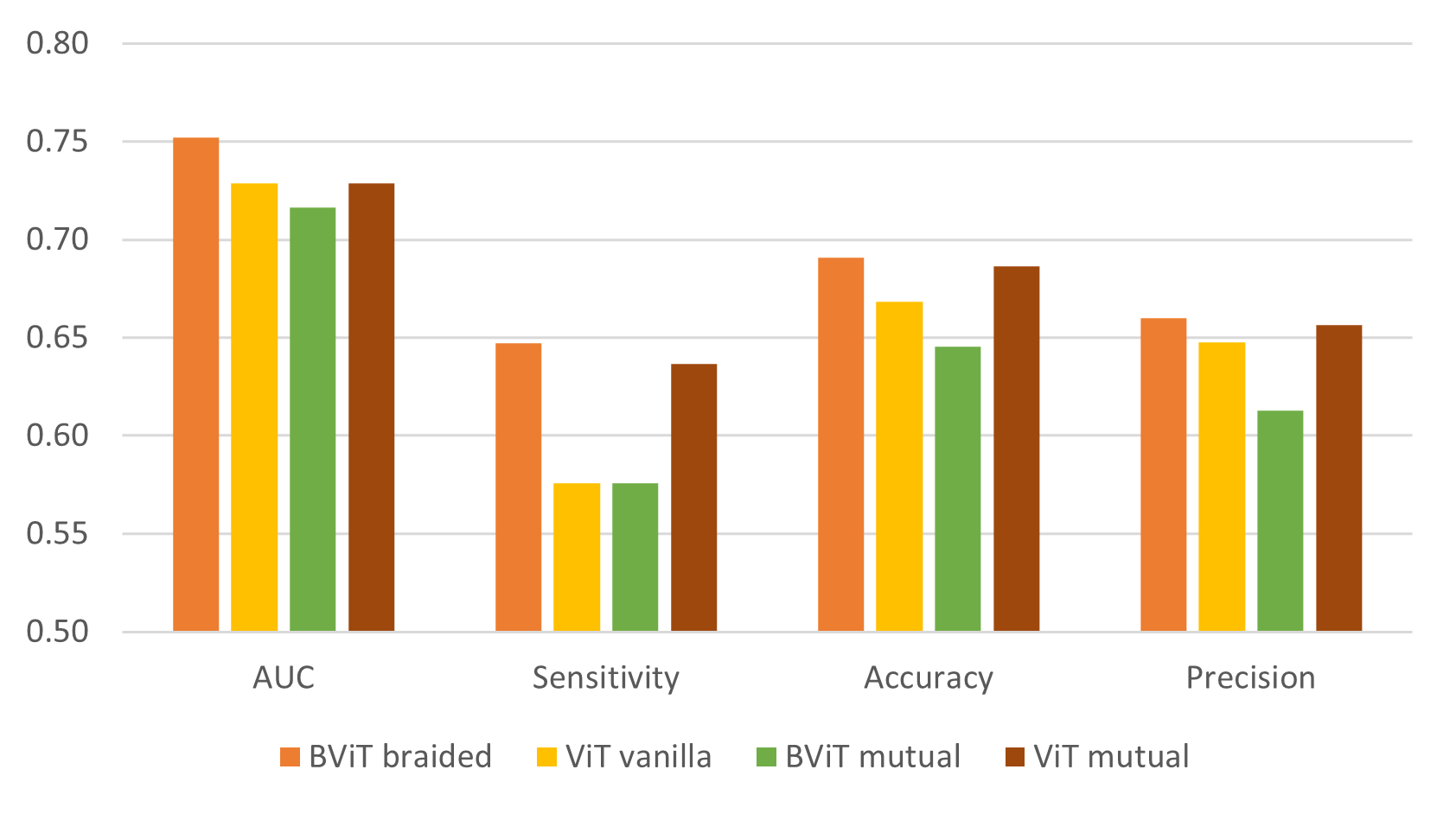}
    \caption{Overall performance comparison of the proposed and baseline models for stroke detection.
    \label{fig:comparison_results}}
\end{figure}

The results for the initial problem formulation, i.e., the discrimination of healthy controls from the combined stroke and TIA groups, are shown in Table \ref{tab:experiments_1}, where BViT achieved the highest area under the curve (AUC) of $0.75$ and sensitivity of $0.65$ for stroke detection. The performance comparison of vision transformers in Fig. \ref{fig:comparison_results} further demonstrates that the proposed BViT with braided transformer module outperforms all the other models, followed by the ViT mutual transformer architecture \cite{wu2022multimodal}. Although the mutual transformers introduce partial cross-channel interactions within their module, incorporating the mutual connection into BViT degrades its performance to the lowest among the other models. This performance drop can be due to the additional architectural complexity, which may hinder effective feature learning rather than improving it.

Table \ref{tab:experiments_2} presents the results for the multi-class discrimination of stroke, TIA, and healthy controls. The proposed BViT with the braided transformer module achieves the highest AUC scores for healthy controls ($0.72$) and TIA patients ($0.62$). For the stroke class, BViT holds the second-highest AUC of $0.715$ following the ViT mutual \cite{wu2022multimodal} with $0.719$ AUC. Overall, AUC scores remain around $0.7$ for healthy controls and stroke classes, whereas TIA yields lower performance with AUC values around $0.6$. However, sensitivity drops significantly for the multi-class discrimination, especially for TIA, which is expected, given the clinical difficulty of discriminating TIA from stroke during the early diagnostic phase. Although an optimal threshold value may balance sensitivity/specificity for TIA, a separate validation set was not reserved due to the limited data availability.

\begin{table*}[b!]
\centering
\caption{Average performance results of vision transformers with different transformer modules reported for each class computed from 5-folds, where the highest scores are highlighted in \textbf{bold}.}
\resizebox{\linewidth}{!}{
\begin{tabular}{clccccccc}
\toprule
Class & \multicolumn{1}{l}{Model} & \multicolumn{1}{c}{\begin{tabular}[c]{@{}c@{}}Transformer\\ Module\end{tabular}} & \multicolumn{1}{c}{Sensitivity} & \multicolumn{1}{c}{Specificity} & \multicolumn{1}{c}{Precision} & \multicolumn{1}{c}{F1-Score} & \multicolumn{1}{c}{AUC} & \multicolumn{1}{c}{Accuracy} \\ 
\midrule

 \multirow{6}{*}{Control} & B-ViT & Braided (ours) & $0.736$ & $0.586$ & $0.685$ & $\textbf{0.709}$ & $\textbf{0.716}$ & $\textbf{0.600}$ \\

 & ViT & Vanilla \cite{dosovitskiy2021an} & $\textbf{0.752}$ & $0.475$ & $0.636$ & $0.689$ & $0.708$ & $0.568$ \\

 & B-ViT & Mutual \cite{wu2022multimodal} & $0.736$ & $0.586$ & $0.685$ & $\textbf{0.709}$ & $0.715$ & $\textbf{0.600}$ \\
 
 & ViT & Mutual \cite{wu2022multimodal} & $0.727$ & $0.596$ & $\textbf{0.688}$ & $0.707$ & $0.714$ & $0.595$ \\ 
 
 &  MVS-Net-v1 \cite{degerli2025advancedassessmentstrokeretinal} & - & $0.636$ & $\textbf{0.626}$ & $0.675$ & $0.655$ & $0.675$ &	$0.555$
\\
  
 &  MVS-Net-v2 \cite{degerli2025advancedassessmentstrokeretinal} & - &  $0.694$ & $0.505$ & $0.632$ & $0.661$ & $0.668$ & $0.546$ \\
   
   \hline

 \multirow{6}{*}{Stroke} &  BiT & Braided (ours) & $0.534$ & $0.748$ & $0.513$ & $0.523$ & $0.715$ & $\textbf{0.600}$ \\

 & ViT & Vanilla \cite{dosovitskiy2021an} & $0.438$ & $\textbf{0.796}$ & $\textbf{0.516}$ & $0.474$ & $0.712$ & $0.568$ \\
 
 & B-ViT & Mutual \cite{wu2022multimodal} & $\textbf{0.548}$ & $0.741$ & $0.513$ & $\textbf{0.530}$ & $0.703$ & $\textbf{0.600}$ \\

 & ViT & Mutual \cite{wu2022multimodal} & $0.534$ & $0.748$ & $0.513$ & $0.523$ & $\textbf{0.719}$ & $0.595$ \\ 
 
 & MVS-Net-v1 \cite{degerli2025advancedassessmentstrokeretinal} & - & $0.534$ & $0.708$ & $0.476$ & $0.503$ & $0.651$ & $0.555$
\\
  
  & MVS-Net-v2 \cite{degerli2025advancedassessmentstrokeretinal} & - & $0.438$ & $0.748$ & $0.464$ & $0.451$ & $0.660$ & $0.546$
 \\
   
   \hline

 \multirow{6}{*}{TIA} & BViT & Braided (ours) & $0.154$ & $0.948$ & $\textbf{0.286}$ & $0.200$ & $0.618$ & $\textbf{0.600}$ \\

 & ViT & Vanilla \cite{dosovitskiy2021an}& $0.077$ & $0.933$ & $0.133$ & $0.098$ & $0.583$ & $0.568$ \\

 & B-ViT & Mutual \cite{wu2022multimodal} & $0.115$ & $\textbf{0.954}$ & $0.250$ & $0.158$ & $0.560$ & $\textbf{0.600}$ \\

 & ViT & Mutual \cite{wu2022multimodal}	& $0.154$ & $0.938$ & $0.250$ & $0.190$ & $0.583$ & $0.595$ \\

 & MVS-Net-v1 \cite{degerli2025advancedassessmentstrokeretinal} & - & $\textbf{0.231}$ & $0.907$ & $0.250$ & $\textbf{0.240}$ & $0.672$ & $0.555$ \\

  & MVS-Net-v2 \cite{degerli2025advancedassessmentstrokeretinal} & - & $0.154$ & $0.928$ & $0.222$ & $0.182$ & $\textbf{0.712}$ & $0.546$ \\

\bottomrule
\end{tabular}}
\label{tab:experiments_2}
\end{table*}

\textbf{Comparison to existing studies.} To the best of our knowledge, this study is the first to investigate the potential of vision transformers for stroke and TIA detection using \textit{multi-view} retinal fundus images. In the literature, only one previous study \cite{degerli2025advancedassessmentstrokeretinal} has explored multi-view retinal fundus imaging for stroke detection, where the authors proposed the Multi-View Stroke Network (MVS-Net), which aggregates multi-view features extracted from retinal images using state-of-the-art CNN backbones. To enable a fair comparison between convolutional and transformer-based architectures, we report the performance of MVS-Net-v1 with a ResNet50 backbone \cite{he2016deep} and MVS-Net-v2 with a DenseNet-121 backbone \cite{huang2017densely}. MVS-Net-v1 has a parameter count of $9,83$ million, comparable to that of BViT with $7,97$ million parameters, whereas MVS-Net-v2 represents the largest-capacity model with $ 34,77$ million parameters among the evaluated architectures. As shown in Table~\ref{tab:experiments_1}, BViT achieved a higher AUC than both MVS-Net variants for binary classification. Similarly, the multi-class results in Table~\ref{tab:experiments_2} demonstrate that the transformer-based models achieved higher AUC values for the control and stroke classes, while MVS-Net-v2 yielded the best performance for the TIA class. Despite the lightweight design of the proposed vision transformer, it is capable of matching or surpassing the established CNN-based multi-view approaches. Overall, the results highlight the potential of transformer architectures as a promising alternative for multi-view retinal fundus image analysis and stroke risk assessment.

\begin{figure}[t!]
    \centering
    \includegraphics[width=0.9\linewidth]{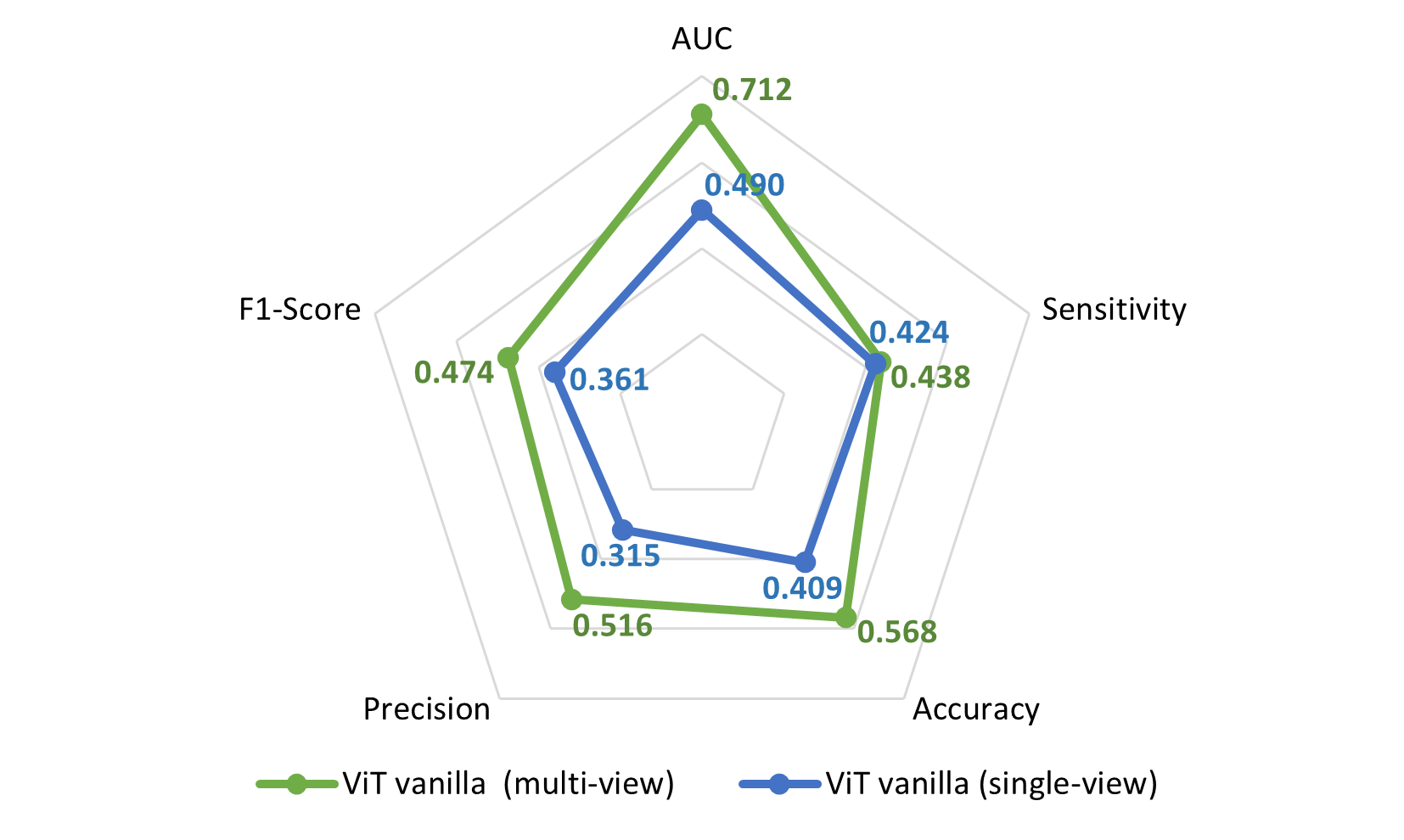}
    \caption{Performance comparison of multi-view and single-view Vision Transformers for stroke detection.
    \label{fig:single_view}}
\end{figure}

Many studies \cite{pachade2022detection, Lim_Lim_Xu_Ting_Wong_Lee_Hsu_2019, coronado2021towards, diagnostics12071714} have focused on stroke detection using only \textit{single-view} retinal fundus images. To assess this setting, we trained a baseline ViT by treating each image independently under the same experimental setup as the multi-view approach. As illustrated in Fig.~\ref{fig:single_view}, the single-view model achieved an AUC of $0.49$, whereas the multi-view version reached $0.71$. This notable improvement indicates that a single retinal image may not provide sufficient information to capture the heterogeneous and spatially distributed retinal biomarkers associated with stroke risk. In contrast, the multi-view approach leverages complementary vascular and structural information from different retinal regions and both eyes, enabling a more comprehensive characterization of disease-related patterns. The substantial improvement from the single-view setting to multi-view highlights the importance of aggregating information across views and suggests that ViTs are particularly effective at modeling global contextual relationships and cross-view information fusion that may not be captured from individual fundus images alone. These findings establish a promising foundation for future transformer-based multi-view retinal imaging frameworks for ischemic stroke prediction.

\textbf{Ablation study.} The vanilla ViT \cite{dosovitskiy2021an} serves as the baseline for the ablation study, as its input channels are processed independently without explicit inter‑channel interactions compared to the braided structure in BViT. As shown in Fig. \ref{fig:comparison_results}, the proposed multi-view modeling outperforms single-view in the vanilla ViT \cite{dosovitskiy2021an} across the presented metrics. These results provide a clear proof‑of‑concept for the effectiveness of the proposed braided architecture in modeling cross‑channel dependencies in retinal fundus imaging.

\section{Conclusion}
This paper investigates stroke assessment from retinal fundus images by proposing a transformer-based architecture, BViT, which jointly analyzes macula-centric and optic nerve head–centric views through a braided design capturing inter-view relationships across both eyes. To the best of our knowledge, this is among the first applications of ViTs for retinal fundus imaging–based stroke assessment. Unlike prior single-view approaches, BViT leverages multi-view information and outperforms vanilla and mutual ViT baselines, achieving an AUC of $0.75$. The proposed approach is suitable for integration into fundus imaging systems, although neuroimaging remains essential for therapeutic decision-making.

% TODO FINAL: This \clearpage needs to be removed from both review and camera-ready versions.

\section*{Acknowledgements}
This study was supported by the Stroke-Data project under Business Finland Grant 3617/31/2019.

\bibliographystyle{splncs04}
\bibliography{main}
\end{document}